\documentclass[pmlr]{jmlr}

\usepackage{longtable}

\usepackage{booktabs}
\usepackage[load-configurations=version-1]{siunitx} 
 \usepackage{graphicx}
\usepackage{amsmath}
\usepackage{threeparttable}
\usepackage{booktabs} 
\usepackage{tabularx}
\usepackage{multirow}
\usepackage{longtable}
\usepackage{ltxtable}
\usepackage{comment}
\usepackage{mathtools}
\usepackage{makecell}
\usepackage{algorithm}
\usepackage{algorithmic}
\usepackage{hyperref}
\usepackage{xcolor}
\usepackage[thinc]{esdiff}
\usepackage{graphicx}
\usepackage{hyperref}

\makeatletter
\def\set@curr@file#1{\def\@curr@file{#1}} 
\makeatother

\theorembodyfont{\upshape}
\theoremheaderfont{\scshape}
\theorempostheader{:}
\theoremsep{\newline}

\jmlrworkshop{Machine Learning for Healthcare}

\title[Short Title]{Bootstrapping Your Own Positive Sample: \\ Contrastive Learning With Electronic Health Record Data}

\author{\Name{Tingyi Wanyan}
       \Email{tiwanyan@iu.edu}\\
       \addr  Indiana University\\ Bloomington,IN,USA
       \AND
       \Name{Jing Zhang}
       \Email{zhang-jing@ruc.edu.cn}\\
       \addr  Renmin University of China
       \AND
       \Name{Ying Ding}
       \Email{ying.ding@ischool.utexas.edu}\\
       \addr   University of Texas at Austin\\
       Austin,Texas,USA
       \AND
       \Name{Ariful Azad}
       \Email{azad@iu.edu}\\
       \addr Indiana University\\
       Bloomington,IN,USA
       \AND
       \Name{Zhangyang Wang}
       \Email{atlaswang@utexas.edu}\\
       \addr University of Texas at Austin\\
       Austin,Texas,USA
       \AND
       \Name{Benjamin S Glicksberg}
       \Email{ benjamin.glicksberg@mssm.edu}\\
       \addr  Icahn School of Medicine at Mount Sinai\\
       NewYork,NY,USA}

\begin{document}

\maketitle

\begin{abstract}
Electronic Health Record (EHR) data has been of tremendous utility in Artificial Intelligence (AI) for healthcare such as predicting future clinical events. These tasks, however, often come with many challenges when using classical machine learning models due to a myriad of factors including class imbalance and data heterogeneity (i.e., the complex intra-class variances). To address some of these research gaps, this paper leverages the exciting contrastive learning framework and proposes a novel contrastive regularized clinical classification model. The contrastive loss is found to substantially augment EHR-based prediction: it effectively characterizes the similar/dissimilar patterns (by its "push-and-pull" form), meanwhile mitigating the highly skewed class distribution by learning more balanced feature spaces (as also echoed by recent findings). In particular, when naively exporting the contrastive learning to the EHR data, one hurdle is in generating positive samples, since EHR data is not as amendable to data augmentation as image data. To this end, we have introduced two unique positive sampling strategies specifically tailored for EHR data: a \textit{feature-based} positive sampling that exploits the feature space neighborhood structure to reinforce the feature learning; and an \textit{attribute-based} positive sampling that incorporates pre-generated patient similarity metrics to define the sample proximity. Both sampling approaches are designed with an awareness of unique high intra-class variance in EHR data. Our overall framework yields highly competitive experimental results in predicting the mortality risk on real-world COVID-19 EHR data with a total of 5,712 patients admitted to a large, urban health system. Specifically, our method reaches a high AUROC prediction score of 0.959, which outperforms other baselines and alternatives: cross-entropy(0.873) and focal loss(0.931). 
\end{abstract}

\section{Introduction}

The use and adoption of electronic health record (EHR) systems in hospitals have rapidly grown in the past decade, and the massive amount of EHR data accumulated from routine care has naturally facilitated a surge of research from data-driven clinical informatics applications, such as medical concept extraction \citep{jiang2011study}, patient trajectory modeling \citep{ebadollahi2010predicting}, disease inference \citep{austin2013using}, and clinical decision support systems \citep{kuperman2007medication}. While primarily designed for operation processes, EHR systems electronically store data associated with each patient encounter with the health system, including disease diagnoses, laboratory test results, vital signs, and more. In recent years, many machine learning techniques, including deep learning, have been leveraged to
derive insights from EHR data \citep{shickel2017deep}.

One of the challenges in learning from EHR data is the heterogeneous nature by which it is represented -- in terms of not only data types involved, but also the various types of contributing factors to disease phenotypes as well as noise, bias, and confounding variables. More specifically, patients with the same disease or outcome can deviate in terms of phenotype representation leading to a high amount of \textbf{intra-class variance} in  terms  of  etiology  and  presentation, especially for complex diseases. Another technical barrier to successful machine learning implementation of EHR data arises from the severe \textbf{data imbalance} as commonly seen in real-world biomedical data. Many important healthcare events are rare, and the class distribution in EHR data is often highly skewed in having a much higher proportion of the majority, background class (i.e., healthy) than those the outcome of interest, such as rare clinical diseases. These aspects can strongly bias the classifier to miss the rare but critical classes. Additionally, many features have high missingness such as lab test results which are sporadic and only performed in certain clinical scenarios.

Recently, contrastive learning \citep{he2019momentum,chen2020simple} has garnered a significant amount of interest, mainly due to its encouraging promise of learning representations while requiring no human annotation that are on par with supervised learning. Originating from the unsupervised learning realm, contrastive learning has also been extended to the semi-supervised \citep{chen2020big} and fully-supervised \citep{khosla2020supervised} settings, allowing for effectively leveraging label information when available. The underlying idea across these settings is to pull ``similar points"  (or points belonging to the same class) together,  while simultaneously pushing apart ``dissimilar points" (or points belonging to different classes), in the embedding space. Accordingly, each training sample is an \textit{anchor} with its similar and dissimilar points in the same training serving as \textit{positive} and \textit{negative} samples, respectively. In another use case \citep{khosla2020supervised}, the supervised variant of contrastive loss was found to consistently perform better than cross-entropy on large-scale classification problems, while also yielding superior robustness to data noise and unseen corruptions during testing. Lately, it was found by \citep{yang2020rethinking,kang2021exploring} that when the data distribution is skewed, contrastive learning can learn a more balanced feature space than its vanilla supervised counterpart.

The above progress sheds light on a new opportunity to integrate contrastive learning into EHR classification tasks, with the hope for stronger discriminative power, robustness, and handling outcome imbalance. However, the incorporation of contrastive learning into EHR data faces one unique roadblock: the process of sampling patients with \textit{positive} outcomes. For any "anchor" sample, a good positive sample has to be semantically aligned (e.g., the same class), yet also nontrivial and informative enough for learning meaningful features. It is well-known that the quality of positive sampling determines the effectiveness of contrastive learning to a large extent \citep{grill2020bootstrap,li2020self}. Previous literature either (in the unsupervised setting) performs data augmentation to create a similar anchor and positive examples \citep{he2019momentum,chen2020simple}, or (in the supervised setting) randomly samples examples from the same class \citep{khosla2020supervised,kang2021exploring}. Unfortunately, EHR data is not as readily amendable to data augmentation due to the previously mentioned aspects (e.g., varying feature completeness). Furthermore, random class-wise sampling overlooks the ultra-high intra-class variance in certain clinical phenotypes and will easily collapse the learned features, which we find in our experiments. In summary, no off-the-shelf positive sampling is directly applicable for contrastive learning in EHR data. 
 
 In this work, we address the above challenges by presenting a holistic framework for classification tasks in EHR data using contrastive learning to explicitly take outcome imbalance and intra-class heterogeneity into consideration. We introduce contrastive loss into a focal loss-based classification pipeline and show that contrastive loss boosts both overall task accuracy in different scenarios of class imbalance in EHR data. As the key innovation, we design two unique positive sampling strategies specifically tailored for EHR data which is less amenable to data augmentation:  a \textit{feature-based} positive sampling that exploits the feature space neighborhood structure to reinforce the feature learning; and an \textit{attribute-based} positive sampling that incorporates pre-generated patient similarity to define the sample proximity. Both sampling approaches are designed to capture the high intra-class heterogeneity in EHR data. Our contributions can be summarized in the following three-folds:
\begin{itemize}
    \setlength\itemsep{0em}
    \item \textbf{Framework:} We are the first to integrate a bespoke approach using contrastive learning into the challenging task of outcome classification for real-world, imbalanced, and heterogeneous EHR data. When combined with a strong baseline using focal loss, we demonstrate that integrating contrastive learning can further remarkably boost predictive accuracy and be robust to data imbalance.
    
    \item \textbf{Methodology:} We present two new positive sampling approaches that enable the usage of contrastive learning in EHR data. Both approaches take better care of the high intra-class variance in EHR data, and outperform existing vanilla options. These approaches may open a new set of possibilities for extending contrastive learning to many other domains where data augmentation is less feasible.
    
    \item \textbf{Experiments:} We test our model on predicting 24-hour mortality using real-world COVID-19 EHR data from a large, diverse health system. We assess our contrastive regularizer and two positive sampling strategies. We assess the robustness of this framework in chunks of data with different sample sizes and imbalance ratios. Our method, particularly the attribute-based positive sampling contrastive regularizer, achieves a boost in performance over vanilla focal loss, reaching a high AUROC prediction score of 0.959, largely outperforming other alternatives. 
\end{itemize}

\section{Related Work}

\textbf{Phenotype Intra-class Heterogeneity and Subphenotypes}. Clinical phenotypes, and therefore EHR data, are often heterogeneous by nature. Most complex diseases, for instance, have varying manifestations, presentations, sequelae, and outcomes. As such, these diseases are manifested by a variety of clinical data types, including lab test result ranges or disease diagnostic codes. Therefore, it is often the case that patients in the same outcome class (i.e., those that develop severe COVID-19) could have large intra-class variance in terms of etiology and presentation. Patterns in this phenomenon can be considered subphenotypes of a disease. Exploring different subphenotypes is valuable to precision medicine and can enhance the performance of the predictive tasks and lead to more personalized recommendations. There is a large body of work exploring computational methods for subphenotyping, such as Parkinson’s disease \citep{Lewis343}, scleroderma \citep{schulam2015clustering}, and Glioblastoma \citep{verhaak2010integrated}.
To better capture the pattern of subtypes, methods such as multi-task learning and hierarchical models  \citep{suresh2018learning,7913647} have been studied. Recently, Su et al. characterized the heterogeneity of COVID-19 into four distinct clinical subphenotypes~\cite{su2021novel}.

\textbf{Contrastive Learning for Data Imbalance}. It is long known that real-world datasets, particularly EHR data, have issues with outcome imbalance which limits performance for many analyses~\citep{santiso2019class,wu2010prediction}(see section 3.2 for more details). Traditional methods like ensemble learning \citep{khalilia2011predicting} or re-balancing classes \citep{buda2018systematic} have been utilized for certain tasks in this realm but come with their own set of limitations. The incorporation of focal loss achieved higher accuracy on EHR-based classification tasks \citep{wang2018utility,wang2020feature}. These facets suggest that contrastive learning may serve to better address imbalance in addition to focal loss. Studies found that decoupling the data representation and classifier can lead to better classification for long-tailed datasets \citep{kang2019decoupling}. \cite{yang2020rethinking} proposed either to use a simple pseudo-labeling strategy to alleviate label bias with extra data in a semi-supervised manner, or to abandon labels at the beginning and pre-train classifiers in a self-supervised manner, can both improve the performance class-imbalanced learning. \cite{kang2021exploring} compared self-supervised contrastive learning and supervised methods for long-tailed datasets and found that self-supervised contrastive learning constantly outperforms supervised methods on heavily imbalanced data. Their study shows that representation learned from the self-supervised contrastive loss performs well on both balanced and imbalanced datasets because it can generate a balanced feature space with similar separability for all classes.  

\textbf{Supervised Contrastive Learning}. Self-supervised contrastive learning takes an augmented anchor as the single positive for each anchor without taking the advantage of pre-labeled class information. So, in one batch, images from the same class of the anchor have been treated as the negative samples which can reduce the performance \citep{khosla2020supervised}. Supervised contrastive learning can leverage label information to generate better positive samples that embeddings of objects from the same classes should be similar, while embeddings of objects from different classes should be dissimilar. \cite{khosla2020supervised} proposed the multiple positives per anchor in addition to many negatives and provide a unified loss function which can be viewed as the generalization of both triplet \citep{weinberger2009distance} and N-pair \citep{sohn2016improved} losses. Their loss is less sensitive to hyperparameters, which can provide consistent boosts for accuracy for different datasets, and is robust to natural corruptions. But taking multiple positive samples from the same class in EHR data can be problematic because there are complex intra-class variances in the EHR data which can lead to over-collapsing feature spaces. \cite{kang2021exploring} proposed k-positive contrastive learning to take advantage of supervised contrastive learning and also solve the issue of imbalanced data. The proposed k-positive contrastive learning takes k instances of the same class of the anchor as the positives and demonstrates superior performance over the latest contrastive learning methods on both balanced data and long-tailed data. The proposed k-positive contrastive loss is different from \citep{khosla2020supervised}’s supervised contrastive learning  which uses all the instances from the same class to be the positive pairs and cannot avoid the dominance of large classes in the representation space. While \cite{kang2021exploring}’s k-positive contrastive loss can carefully balance the equal number of positive pairs for all classes, especially for the long-tailed data which class instances vary dramatically. Therefore, k-positive contrastive loss can generate feature spaces with desirable balance and discriminative ability. 

\textbf{Applications of Contrastive Learning on Clinical Data}. There have been encouraging studies that have demonstrated the potential utility of contrastive learning in health data, albeit only a few. Contrastive learning has been applied for more robust learning of various patient data modalities.~\cite{kiyasseh2020clocs} created CLOCS, a family of contrastive learning methods on unlabeled cardiac physiological data for downstream tasks like better quantifying patient similarity for disease detection. \cite{kostas2021bendr} created BENDR, which leverages transformers and contrastive self-supervised learning to better learn representations of electroencephalogram data. Moving to the realm of EHR data, \cite{li2019distributed} built a framework that enhanced predictive performance for common diseases across multiple sites without the need to share data by leveraging Distributed Noise Contrastive Estimation. \cite{wanyan2021contrastive} demonstrated that contrastive learning enhanced prediction of critical events in COVID-19 as well as led to better patient representations. \cite{chen2021disease} used transformers and contrastive learning to learn embedding representations of EHR data and showed that these representations allow for better predictions in disease retrieval tasks. It is clear that the potential utility of this framework into the realm of healthcare, especially EHR, is just at the beginning.

\section{Methodology}
\subsection{Addressing EHR Data Imbalance: Contrastive-Regularized Focal Loss}

Focal loss ~\citep{lin2017focal} has been shown to work well in imbalanced EHR data and significantly improves performance in such tasks as predicting mortality from heart failure~\citep{wang2020feature}. However, focal loss may underperform in situations where there is a lot of patient intra-class heterogeneity. This may be the case because of the unimodal loss structure of focal loss which classifies based solely on label information, making it unable to leverage and learn from the rich features that constitute patient data within a group~\citep{oord2018representation}. To leverage the above problem, we propose a learning framework by adding a contrastive regularizer to the base focal loss, for boosting the performance on EHR tasks on outcome imbalance as well as intra-class heterogeneity. 
We apply our contrastive learning framework similar to~\cite{chen2020simple}, then compile an end-to-end training strategy that incorporates our novel k-positive selection approaches (described in more detail in the next section):

\begin{equation}
    \mathcal{L}=\mathcal{\hat{L}}+\alpha \mathcal{L^{*}}
\end{equation}
Where $\mathcal{\hat{L}}$ is the focal loss, $\mathcal{L^{*}}$ is the additional contrasitve loss as a regularizer, $\alpha$ is the regularization coefficient that controls the loss magnitude. Different from~\citep{chen2020simple}, we use the supervised contrastive learning framework~\citep{khosla2020supervised} to generate augmented positive samples from the same class group, for each batch, we sample N samples, and use our proposed sampling strategies to generate K positive samples from the same class for each data in batch, then use all $(N-1)\times K$ samples  as the negative samples:
\begin{equation}
    \mathcal{L^{*}}=-\frac{1}{N}\sum^{N}_{p=1}\sum^{K}_{i=1}\log\frac{\exp(\vec{z_{p}}\cdot \vec{z_{i}^{+}}/\tau)}{\exp(\vec{z_{p}}\cdot \vec{z_{i}^{+}}/\tau)+\sum^{(N-1)K}_{j=1}\exp(\vec{z_{p}}\cdot \vec{z_{j}^{-}}/\tau)}
    \label{eq:krandsampling}
\end{equation}
Where $\vec{z_{p}}$ is the embedding vector of one patient data in batch, $\vec{z_{i}^{+}}$ are positive sample embeddings. $\vec{z_{j}^{-}}$ are negative sample embeddings. $K$ is the positive sample numbers, $\tau$ is the temperature hyper-parameter. Intuitively, we apply a vanilla k-random positive sampling strategy~\citep{kang2021exploring} as our first positive sampling strategy as well as a baseline for comparison with our next proposed two rational sampling strategies, the learning algorithm with this k-random positive sampling is shown in Alg~\ref{alg}. We highlight the part in red to mainly distinguish the K-random sampling algorithm from our next proposed two positive sampling strategies.

\begin{algorithm}[h]

\caption{K-Random Positive Sampling Contrastive Learning}
\label{alg}
\textbf{Input}: longitudinal EHR features\\
\textbf{Output}: Embedding vector representation for patients
\begin{algorithmic}[1] 
\STATE Initialize all weights $W$
\FOR{each epoch}
\WHILE{not converged}
\STATE sample a mini-batch training patients $P\in P_{all}$.
\STATE \textcolor{red}{for each $p\in P$, randomly sample k positive data $p^{+}_{k}\in P_{all}$ that have the same label as $p$.}
\STATE Compute: $\mathcal{L}=\mathcal{\hat{L}}+\alpha \mathcal{L^{*}}$
\STATE compute gradient of loss function $\nabla \mathcal{L}$ and update weight matrices $W$.
\ENDWHILE
\ENDFOR
\STATE \textbf{return} embedded representation $c_p,\forall{p}\in P_{all}$
\end{algorithmic}
\end{algorithm}

\subsection{Addressing EHR Data Heterogeneity: Two Proximity-based Positive Sampling Approaches}

A key knob in contrastive learning is to find positive pairs for anchor examples and to maximize their learned features' similarity to inject the desired invariance. As class labels are available for us, a vanilla option is to follow the k-random positive sampling strategy~\citep{kang2021exploring} for supervised contrastive learning, which just randomly picks samples belonging to the same class to form anchor-positive pairs. However, we demonstrate this naive baseline is unable to work well for EHR data since it neglects the high intra-class variance as seen in heterogeneous phenotypes like COVID-19, and such learned features will easily collapse and fail to generalize. To this end, we develop two unique positive sampling strategies specifically tailored for EHR data: a \textit{feature-based} positive sampling that exploits the feature space neighborhood structure to reinforce the feature learning; and an \textit{attribute-based} positive sampling that incorporates the raw features to define the sample proximity. Both sampling approaches are designed to capture the high intra-class variance in EHR data. The main difference between the two lies in how they compute sample similarity, which has been highlighted in Algorithm 2 (\textcolor{blue}{blue}) and Algorithm 3 (\textcolor{orange}{orange}), respectively.



\textbf{Feature-based Sampling}
 In our feature-based k nearest neighborhood (knn) positive sampling method, we construct a knn graph by ranking patients by their similarities among the embedding vectors within the same class and selecting positive samples for every training patient from its top k neighbors in the feature knn graph. We define the similarity score between one pair of features as their cosine similarity:
\begin{equation}
    \vec{z}\cdot \vec{\bar{z}}/ \| \vec{z}\| \|\vec{\bar{z}}\|
    \label{feature_sim}
\end{equation}
Where $\vec{z}$ and $\vec{\bar{z}}$ are two embedding feature vectors. Since constructing a knn graph would take large computational resources, especially when data is huge, we update the knn graph every epoch instead of every iteration. Our feature-based knn sampling contrastive regularizer loss is then written as follow:
\begin{equation}
\label{eq:feature}
    \mathcal{L^{*}}_{feature}=-\frac{1}{N}\sum^{N}_{p=1}\sum^{K}_{i=1}\log\frac{\exp(\vec{z_{p}}\cdot \vec{z_{i(feature)}^{+}}/\tau)}{\exp(\vec{z_{p}}\cdot \vec{z_{i(feature)}^{+}}/\tau)+\sum^{(N-1)K}_{j=1}\exp(\vec{z_{p}}\cdot \vec{z_{j}^{-}}/\tau)}
\end{equation}
Where $\vec{z_{i(feature)}^{+}}$ are the top k sample embeddings from the feature knn graph. The knn feature based contrastive learning algorithm is shown in Alg~\ref{alg_knn}. 
\begin{algorithm}[h]
\caption{Feature-based K-nearest Neighborhood Sampling Contrastive Learning}
\label{alg_knn}
\textbf{Input}: Longitutinal EHR features\\
\textbf{Output}: Embedding vector representation for all patients
\begin{algorithmic}[1] 
\STATE Initialize all weights $W_p$
\FOR{each epoch}
\STATE 
\textcolor{blue}{Compute similarities between all pairs of embedding feature representations based on equation~\ref{feature_sim}, and build knn graph from it.}
\WHILE{not converged}
\STATE sample a mini-batch training patients $P\in P_{all}$.
\STATE for each $p\in P$, sample k positive data $p^{+}_{k}\in P_{all}$ that have the same label as $p$, and are connected to node $p$ in the knn graph.
\STATE Compute: $\mathcal{L}=\mathcal{\hat{L}}+\alpha \mathcal{L^{*}}_{feature}$
\STATE compute gradient of loss function $\nabla \mathcal{L}$ and update weight matrices $W_{p}$.
\ENDWHILE
\ENDFOR
\STATE \textbf{return} embedded representation $c_p,\forall{p}\in P_{all}$
\end{algorithmic}
\end{algorithm}

\textbf{Attribute-based Sampling}
In our attribute-based positive sampling model, we construct the knn graph to rank patients by their similarities by lab test features. Since the input values are in different scales, we scale the feature values into the range of [0,1] based on their mean and standard deviation, and define the attribute similarity score by their Euclidean distance. We included 34 lab test features that were present in ~80\% of the cohort (see Appendix A):
\begin{equation}
    \sum^{m}_{\bar{k}=1}\|X_{i\bar{k}}-X_{j\bar{k}}\|_{2}
    \label{attribute_sim}
\end{equation}
Where $X_i$ and $X_j$ are two input vectors representing two individual patients, $m$ is the input feature dimension. $\bar{k}$ represents each highlighted feature. The contrastive regularizer loss is written as follow:
\begin{equation}
\label{eq:attribute}
    \mathcal{L^{*}}_{attribute}=-\frac{1}{N}\sum^{N}_{p=1}\sum^{K}_{i=1}\log\frac{\exp(\vec{z_{p}}\cdot \vec{z_{i(attribute)}^{+}}/\tau)}{\exp(\vec{z_{p}}\cdot \vec{z_{i(attribute)}^{+}}/\tau)+\sum^{(N-1)K}_{j=1}\exp(\vec{z_{p}}\cdot \vec{z_{j}^{-}}/\tau)}
\end{equation}
Where $\vec{z_{i(attribute)}^{+}}$ are the top k sample embeddings from the attribute knn graph. Alg~\ref{alg_knn_att} shows our attribute-based contrastive learning algorithm. While our feature-based sampling strategy updates the knn graph every epoch, the attribute-based knn graph is computed before training begins. In other words, we use pre-computed patient similarity metrics to select positive samples which are not learned by our algorithm. 
\begin{algorithm}[h]
\caption{Attribute Based K-nearest Neighborhood Sampling Contrastive Learning}
\label{alg_knn_att}
\textbf{Input}: Longitudinal EHR features\\
\textbf{Output}: Embedding vector representation for all patients
\begin{algorithmic}[1] 
\STATE Initialize all weights $W_p$
\STATE \textcolor{orange}{Pre-compute similarities between all pairs of embedding features based on equation~\ref{attribute_sim}, and build knn graph from it}.
\FOR{each epoch}
\WHILE{not converged}
\STATE sample a mini-batch training patients $P\in P_{all}$.
\STATE for each $p\in P$, sample k positive data $p^{+}_{k}\in P_{all}$ that have the same label as $p$, and are connected to node $p$ in the knn graph.
\STATE Compute: $\mathcal{L}=\mathcal{\hat{L}}+\alpha \mathcal{L^{*}}_{attribute}$
\STATE compute gradient of loss function $\nabla \mathcal{L}$ and update weight matrices $W_{p}$.
\ENDWHILE
\ENDFOR
\STATE \textbf{return} embedded representation $c_p,\forall{p}\in P_{all}$
\end{algorithmic}
\end{algorithm}
\section{Experiment}
\subsection{Dataset Description and Experimental Design}
We assess our framework in a clinically-relevant EHR task, specifically predicting mortality from real-world COVID-19 data. Our dataset is comprised of patients from a large and diverse health system in an urban environment. We obtain such data for 5,712 patients who tested positive for COVID-19 and were hospitalized (~23\% mortality rate). The collected EHR data contains the following information: COVID-19 status, demographics, laboratory test results, vital signs, and comorbidities (see Appendix A for details).  

Our primary task was to predict mortality in COVID-19 patients 24 hours before the event. We model the EHR data using a standard longitudinal RNN model framework as in~\cite{choi2016doctor}. Longitudinal data (i.e., features with multiple values), specifically lab tests and vital signs, were binned and averaged within 6-hour windows across their hospitalization. We concatenated non-longitudinal categorical features, specifically demographics and co-morbidities, into a separate shallow neural network layer. We then concatenated the output embedding vector from this layer with the embedding vector from the RNN model (i.e., the last time frame) to form the final patient embedding representation $\vec{z_{p}}$. All analyses were performed using TensorFlow 1.15.1 and utilized the Adam optimizer~\citep{kingma2014adam}. We set the batch size to be 32, with 6 training epochs, and set our embedding dimension to be 100. For all experiment below, we specifically pick our model parameters to be: $k=5$, $\alpha=0.2$, and $\tau=1$. We conduct a thorough investigation of optimal model parameters in Appendix B.

\subsection{Performance of Contrastive Regularizer Positive Sampling Strategies}

For testing the overall performance of our contrastive regularizer on our task, we split our data into 70\% for training and 30\% for testing for performing the 7-fold cross-validation. The performance metrics are shown in Table~\ref{tab:overall}. We first assess the effect of focal loss compared to the commonly used cross-entropy loss. In this comparison, we find focal loss outperforms cross-entropy loss in terms of both AUROC (~6\% improvement) and AUPRC (~3\% improvement).

We then assess any performance improvements over the base focal loss for our various positive sampling approaches, specifically k-random (random), feature-based (feature), and attribute-based (attribute). We find that all of our three positive sampling strategies confer performance improvements over the base focal loss, specifically ~2\%, ~3\%, and ~3\% for random, feature, and attribute respectively. We then compare the overall contribution of the two sampling approaches we developed (feature and attribute) compared to the random k-positive selection for focal loss.

We further assess the relative benefit to these sampling approaches in different EHR cohort scenarios in subsequent generalizability experiments described below.

\begin{table}[h]
  \centering
  \caption{Overall Mortality Prediction Performance}
    \begin{tabular}{p{3cm}p{2cm}p{2cm}}
    \toprule
    Loss Model &AUROC &AUPRC \\
    \midrule
    CE &0.873(0.003) &0.801(0.002)\\
    FL &0.931(0.002) &0.830(0.002)\\
    FL(random) &0.949(0.002) &0.879(0.001) \\
    FL(feature) &0.956(0.002) &0.884(0.001)\\
    FL(attribute) &\textbf{0.959}(0.002) &\textbf{0.886}(0.001)\\
    \midrule
    \end{tabular}%
  \label{tab:overall}%
\end{table}

\subsection{Assessing the Robustness of the Contrastive Regulatrizer Framework with Different Training Sample Sizes and Imbalance Ratios}

Next, we test the robustness of our model and baselines to predict the same outcome in subsets of our data at different sample sizes and imbalance ratios in the training data. Specifically, we train on varying dataset characteristics (i.e. size and imbalance ratio) and test in the fixed original size and imbalance ratio as the original experiment (N=1713). Like before, we assess various baselines and implementations of positive sampling strategies. 

First, the AUROC results of varying training sample sizes are shown in Table~\ref{tab:train_vary}. Specifically, we compared performance at N=399, 999, 1999, 2999, and 3999 all with the same imbalance ratio of 23\% positive outcomes. Across all scenarios, we see a marked improvement of focal loss over cross-entropy loss with the larger improvements seen at smaller sample sizes (i.e.,~7\% improvement in N=1999). The addition of random positive sampling for focal loss increased performance across all experiments but was most prominent in the smallest sample size, specifically~8\% improvement over base focal loss. The incorporation of feature- and attribute-based sampling strategies also demonstrated improvements over the random sampling strategy in focal loss:~1.1\% for feature-based and~2.0\% for attribute-based. For these experiments, we did not find a large difference between the feature-based and attribute-based strategies with slightly higher values in attribute-based.

\begin{table}[t]
  \centering
  \caption{Prediction Performance on Various Sample Sizes in the same Test Set}
    \begin{tabular}{p{2cm}p{2cm}p{2cm}p{2cm}p{2cm}p{2cm}}
    \toprule
    & \multicolumn{4}{l}{Training data(size/positive label percentage)}\\
    \midrule 
    Loss Model &399/23\% &999/23\% &1999/23\% & 2999/23\% &3999/23\%\\
    \midrule
    CE &0.732(0.003) &0.845(0.002) &0.863(0.002) &0.876(0.002) &0.887(0.002)\\
    FL &0.803(0.002) &0.916(0.002) &0.926(0.002) &0.934(0.002) &0.935(0.001)\\
    FL(random)&0.886(0.001) &0.931(0.002) &0.941(0.001) &0.947(0.001) &0.949(0.001)\\
    FL(feature)&0.897(0.001) &0.937(0.001) &0.944(0.001) &0.952(0.001) &\textbf{0.956}(0.001)\\
    FL(attribute)&\textbf{0.906}(0.001) &\textbf{0.938}(0.001) &\textbf{0.949}(0.001) &\textbf{0.953}(0.001) &0.954(0.002)\\
    \midrule
    \end{tabular}%
  \label{tab:train_vary}%
\end{table}

\begin{table}[h]
  \centering
  \caption{Prediction Performance When Varying Training Outcome Imbalance Ratios}
    \begin{tabular}{p{3cm}p{2cm}p{2cm}p{2cm}p{2cm}p{2cm}}
    \midrule
     & \multicolumn{4}{l}{Training data(size/positive label percentage)}\\
     \midrule
    Loss Model &3099/1\% &3245/5\% &3445/10\% & 3645/15\% &3845/20\%\\
    \midrule
    CE &0.693(0.002) &0.801(0.002) &0.828(0.002) &0.841(0.002) &0.847(0.002)\\
    FL &0.849(0.002) &0.884(0.002) &0.901(0.002) &0.927(0.002) &0.932(0.002)\\
    FL(random)&0.929(0.001) &0.934(0.001) &0.939(0.001) &0.944(0.001) &0.949(0.001)\\
    FL(feature)&0.934(0.001) &0.942(0.001) &0.944(0.001) &0.943(0.001) &0.953(0.002)\\
    FL(attribute)&\textbf{0.940}(0.002) &\textbf{0.942}(0.001) &\textbf{0.948}(0.001) &\textbf{0.949}(0.001) &\textbf{0.959}(0.001)\\
    \midrule
    \end{tabular}%
  \label{tab:train_vary_imbalance}%
\end{table}
We next performed a similar experiment varying degrees of data imbalance via restricting the number of positive outcome patients (same amount of negative) in the training data, specifically at 1\%, 5\%, 10\%, 15\%, and 20\%.  Table~\ref{tab:train_vary_imbalance} shows the AUROC results for this experiment. Like before, focal loss has large improvements over cross-entropy loss, with the largest improvement at the most imbalanced ratio, specifically ~15\% improvement at 1\% imbalance. All contrastive regularizers with different positive sampling greatly boost the performance of focal loss, specifically ~9\% improvement. All three positive sampling strategies conferred improvements over the baseline focal loss with the trend of showing bigger improvements with higher training imbalance. The feature- and attribute-based  approaches still show improvements over random sampling across different imbalance ratios, with an average ~0.07\% for feature-based and ~0.11\% for attribute-based. The attribute-based sampling has slight improvement over feature-based sampling, one average~0.04\% at different ratios, note that all contrastive regularizer boosted focal losses have much stable performance at different ratios, with the largest degradation of~2\% from 20\% ratio to 1\% ratio, comparing to the degradation of 15\% from cross-entropy loss and 8.3\% from focal loss, our contrastive regularizer boosted focal loss provides much stable performance.

\subsection{The Impact of Different Positive Sampling Strategies on Representations Learned from EHR Data}
\begin{figure*}
        \centering
        \includegraphics[width=\textwidth]{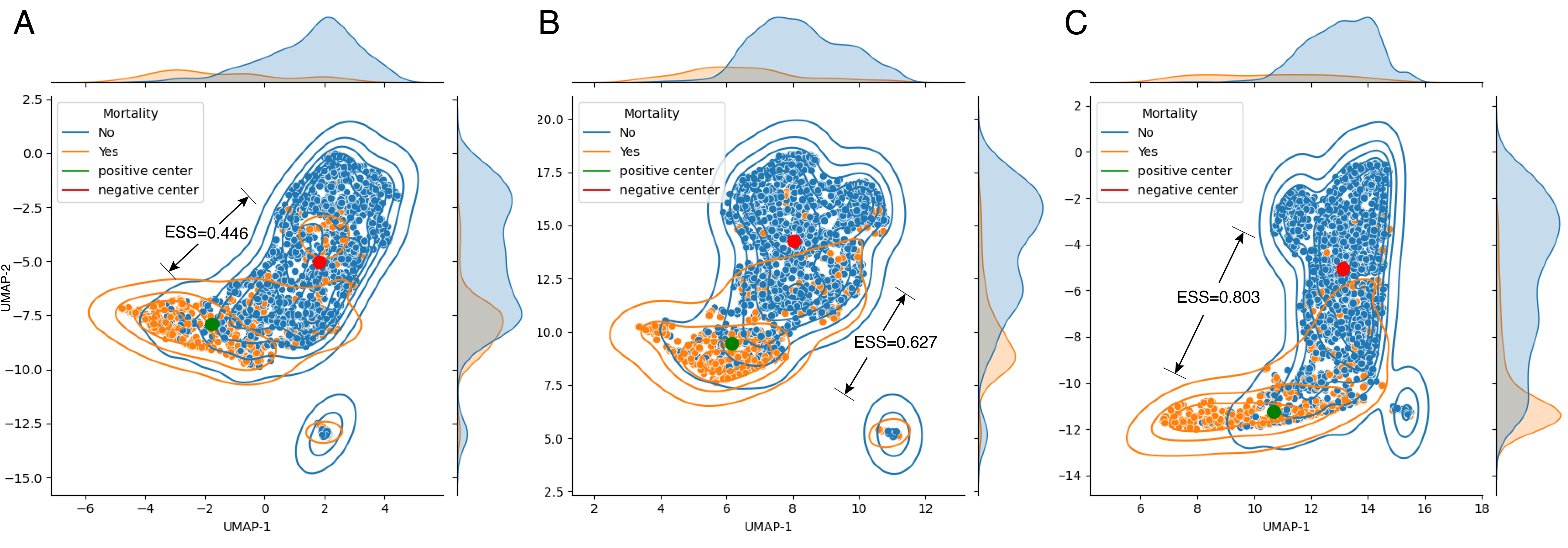}
        \centering
        \vspace{-15pt}
        \caption{UMAP visualizations of pre-trained embedding with different sampling strategies: (A) random positive sampling; (B) feature-based positive sampling; (C) attribute-based positive sampline. Blue dots are non-death patient embeddings, orange dots are death patient embeddings, the orange and blue lines represent distribution contour of the positive and negative groups. \vspace{-15pt}}
        \label{Embeddings}
\end{figure*}

In all previous experiments, we used the contrastive-regularized focal loss in a semi-supervised setting to directly classify patients from the trained model.
Table~\ref{tab:overall} already demonstrated the benefit of our contrastive regularizer with an improved predictive performance in the semi-supervised setting.
Here, we demonstrate the impact of different sampling strategies on the hidden representations learned via supervised contrastive learning~\citep{khosla2020supervised}.
In this setting, we learn 100-dimension hidden representation of patients using the contrasting loss functions shown in Eq.~\ref{eq:krandsampling},~\ref{eq:feature}, and ~\ref{eq:attribute}.
Fig.~\ref{Embeddings} shows the embeddings obtained from different sampling strategies.

\begin{table}[h]
  \centering
  \caption{Pre-trained Embedding Evaluation for Sampling Strategies}
    \begin{tabular}{p{3.5cm}p{2cm}p{2cm}p{2.5cm}p{3cm}}
    \midrule 
    Sampling Strategy  &AUROC &ESS &SD (positive) &SD (negative) \\
    \midrule
    Random&0.859 &0.446 &0.158 &0.448\\
    Feature&0.893 &0.627 &0.193 &0.534\\
    Attribute&\textbf{0.904} &\textbf{0.803} &\textbf{0.255} &\textbf{0.702}\\
    \midrule
    \end{tabular}%
  \label{tab:linear_evl}%
\end{table}

{\bf Feature- and attribute-based positive samplings better separate classes in embedding space.}
To test the separation of classes in embedding space, we train a logistic linear classifier on patient embeddings obtained from contrastive learning with different sampling strategies.
To train the logistic regression model, we use the same split percentage for training and testing data as used in the above section. The first column of Table~\ref{tab:linear_evl} shows that the prediction scores of the linear classifier trained on both attribute- and feature-based embedding outperform the random sampling strategy, specifically 3.4\% for feature-based, and 4.5\% for attribute-based. This result demonstrates that feature- and attribute-based samplings push positive and negative classes further apart in embedding space, which helps the linear classifier attain better accuracy. 

To quantitatively measure the separation between the positive and negative patient groups, we define a simple inter-class distance metric (here we use Embedding Separation Score (ESS) for this metric) as follows: 
\begin{equation}
    \|\vec{z_p}- \vec{\vec{z_n}}\|/ (\| \vec{z_p}\|+\|\vec{\vec{z_n}}\|),
    \label{ESS}
\end{equation}
where $\vec{z_p}$ and $\vec{z_n}$ are normalized embedding centers for positive patient group and negative patient group.
Here, the value of inter-class distance is between $[0,1]$, and the higher the distance the better the classes are separated in the embedding space. Table~\ref{tab:linear_evl} column 2 shows that feature-based sampling and attribute-based sampling show higher inter-class distance, as expected.
Fig.~\ref{Embeddings} visually confirms that the new sampling strategies indeed better separate positive and negative classes. 

{\bf Feature- and attribute-based positive samplings capture more intra-class heterogeneity.}
Usually, it is desirable that points from the same class are pulled together in embedding space to create a compact cluster. 
However, COVID EHR data shows tremendous heterogeneity in terms of patients demographics, symptoms, and outcomes~\citep{su2021novel}. 
Hence, it is often beneficial to keep some intra-class variability (heterogeneity) to facilitate 
learning from local neighborhoods of patients across the spectrum within each class. 
We quantitatively measure the intra-class variance (in the embedding space) to represent intra-class heterogeneous by computing the standard deviation of distance metric between patient embedding with its cluster center embedding in terms of Eq.~\ref{ESS} for the positive class (mortality) patient group and negative patient group, respectively.
Columns 4 and 5 in Table~\ref{tab:linear_evl} show that feature-based and attribute-based sampling indeed preserve higher intra-class variances to reflect the heterogeneity within each class. This benefit is because random positive sampling contrastive loss approximates one uniform distribution within each group, resulting in collapsing the embedding space, while our feature- or attribute-based positive sampling contrastive loss strategies approximate different subgroup distributions within the same class and avoid collapsing of the embedding space. Such collapsing could result in a much closer distribution center between the positive and negative group as shown in Table~\ref{tab:linear_evl}, thus leading to worse prediction performance.

\section{Conclusion}

In this work, we introduce a general framework that adds a contrastive regularizer on top of focal loss for boosting predictive performance. We further propose two novel positive sampling strategies, feature-based and attribute-based, that outperform k random sampling for contrastive learning especially in datasets with high intra-class heterogeneity and data imbalance. Through experiments predicting mortality in real-world COVID-19 EHR data, we demonstrate the contrastive regularized framework greatly boosts the performance over focal loss at various sample sizes and imbalance ratios. Our results show that the two sampling strategies both outperform k random sampling in this task with the attribute-based approach having a slight edge over the feature-based approach. Our experiments further confirm that the two proposed sampling strategies in our contrastive regularized framework can achieve better inter-class separation and leverage intra-class heterogeneity. 


\bibliography{ref}

\begin{thebibliography}{37}
\providecommand{\natexlab}[1]{#1}
\providecommand{\url}[1]{\texttt{#1}}
\expandafter\ifx\csname urlstyle\endcsname\relax
  \providecommand{\doi}[1]{doi: #1}\else
  \providecommand{\doi}{doi: \begingroup \urlstyle{rm}\Url}\fi

\bibitem[{Alaa} et~al.(2018){Alaa}, {Yoon}, {Hu}, and {van der
  Schaar}]{7913647}
A.~M. {Alaa}, J.~{Yoon}, S.~{Hu}, and M.~{van der Schaar}.
\newblock Personalized risk scoring for critical care prognosis using mixtures
  of gaussian processes.
\newblock \emph{IEEE Transactions on Biomedical Engineering}, 65\penalty0
  (1):\penalty0 207--218, 2018.
\newblock \doi{10.1109/TBME.2017.2698602}.

\bibitem[Austin et~al.(2013)Austin, Tu, Ho, Levy, and Lee]{austin2013using}
Peter~C Austin, Jack~V Tu, Jennifer~E Ho, Daniel Levy, and Douglas~S Lee.
\newblock Using methods from the data-mining and machine-learning literature
  for disease classification and prediction: a case study examining
  classification of heart failure subtypes.
\newblock \emph{Journal of clinical epidemiology}, 66\penalty0 (4):\penalty0
  398--407, 2013.

\bibitem[Buda et~al.(2018)Buda, Maki, and Mazurowski]{buda2018systematic}
Mateusz Buda, Atsuto Maki, and Maciej~A Mazurowski.
\newblock A systematic study of the class imbalance problem in convolutional
  neural networks.
\newblock \emph{Neural Networks}, 106:\penalty0 249--259, 2018.

\bibitem[Chen et~al.(2020{\natexlab{a}})Chen, Kornblith, Norouzi, and
  Hinton]{chen2020simple}
Ting Chen, Simon Kornblith, Mohammad Norouzi, and Geoffrey Hinton.
\newblock A simple framework for contrastive learning of visual
  representations.
\newblock In \emph{International conference on machine learning}, pages
  1597--1607. PMLR, 2020{\natexlab{a}}.

\bibitem[Chen et~al.(2020{\natexlab{b}})Chen, Kornblith, Swersky, Norouzi, and
  Hinton]{chen2020big}
Ting Chen, Simon Kornblith, Kevin Swersky, Mohammad Norouzi, and Geoffrey
  Hinton.
\newblock Big self-supervised models are strong semi-supervised learners.
\newblock \emph{arXiv preprint arXiv:2006.10029}, 2020{\natexlab{b}}.

\bibitem[Chen et~al.(2021)Chen, Lo, Lai, and Huang]{chen2021disease}
Yen-Pin Chen, Yuan-Hsun Lo, Feipei Lai, and Chien-Hua Huang.
\newblock Disease concept-embedding based on the self-supervised method for
  medical information extraction from electronic health records and disease
  retrieval: Algorithm development and validation study.
\newblock \emph{Journal of Medical Internet Research}, 23\penalty0
  (1):\penalty0 e25113, 2021.

\bibitem[Choi et~al.(2016)Choi, Bahadori, Schuetz, Stewart, and
  Sun]{choi2016doctor}
Edward Choi, Mohammad~Taha Bahadori, Andy Schuetz, Walter~F Stewart, and Jimeng
  Sun.
\newblock Doctor ai: Predicting clinical events via recurrent neural networks.
\newblock In \emph{Machine learning for healthcare conference}, pages 301--318.
  PMLR, 2016.

\bibitem[Ebadollahi et~al.(2010)Ebadollahi, Sun, Gotz, Hu, Sow, and
  Neti]{ebadollahi2010predicting}
Shahram Ebadollahi, Jimeng Sun, David Gotz, Jianying Hu, Daby Sow, and
  Chalapathy Neti.
\newblock Predicting patient’s trajectory of physiological data using
  temporal trends in similar patients: a system for near-term prognostics.
\newblock In \emph{AMIA annual symposium proceedings}, volume 2010, page 192.
  American Medical Informatics Association, 2010.

\bibitem[Grill et~al.(2020)Grill, Strub, Altch{\'e}, Tallec, Richemond,
  Buchatskaya, Doersch, Pires, Guo, Azar, et~al.]{grill2020bootstrap}
Jean-Bastien Grill, Florian Strub, Florent Altch{\'e}, Corentin Tallec,
  Pierre~H Richemond, Elena Buchatskaya, Carl Doersch, Bernardo~Avila Pires,
  Zhaohan~Daniel Guo, Mohammad~Gheshlaghi Azar, et~al.
\newblock Bootstrap your own latent: A new approach to self-supervised
  learning.
\newblock \emph{arXiv preprint arXiv:2006.07733}, 2020.

\bibitem[He et~al.(2019)He, Fan, Wu, Xie, and Girshick]{he2019momentum}
Kaiming He, Haoqi Fan, Yuxin Wu, Saining Xie, and Ross Girshick.
\newblock Momentum contrast for unsupervised visual representation learning.
\newblock \emph{arXiv preprint arXiv:1911.05722}, 2019.

\bibitem[Jiang et~al.(2011)Jiang, Chen, Liu, Rosenbloom, Mani, Denny, and
  Xu]{jiang2011study}
Min Jiang, Yukun Chen, Mei Liu, S~Trent Rosenbloom, Subramani Mani, Joshua~C
  Denny, and Hua Xu.
\newblock A study of machine-learning-based approaches to extract clinical
  entities and their assertions from discharge summaries.
\newblock \emph{Journal of the American Medical Informatics Association},
  18\penalty0 (5):\penalty0 601--606, 2011.

\bibitem[Kang et~al.(2019)Kang, Xie, Rohrbach, Yan, Gordo, Feng, and
  Kalantidis]{kang2019decoupling}
Bingyi Kang, Saining Xie, Marcus Rohrbach, Zhicheng Yan, Albert Gordo, Jiashi
  Feng, and Yannis Kalantidis.
\newblock Decoupling representation and classifier for long-tailed recognition.
\newblock \emph{arXiv preprint arXiv:1910.09217}, 2019.

\bibitem[Kang et~al.(2021)Kang, Li, Xie, Yuan, and Feng]{kang2021exploring}
Bingyi Kang, Yu~Li, Sa~Xie, Zehuan Yuan, and Jiashi Feng.
\newblock Exploring balanced feature spaces for representation learning.
\newblock In \emph{International Conference on Learning Representations}, 2021.

\bibitem[Khalilia et~al.(2011)Khalilia, Chakraborty, and
  Popescu]{khalilia2011predicting}
Mohammed Khalilia, Sounak Chakraborty, and Mihail Popescu.
\newblock Predicting disease risks from highly imbalanced data using random
  forest.
\newblock \emph{BMC medical informatics and decision making}, 11\penalty0
  (1):\penalty0 1--13, 2011.

\bibitem[Khosla et~al.(2020)Khosla, Teterwak, Wang, Sarna, Tian, Isola,
  Maschinot, Liu, and Krishnan]{khosla2020supervised}
Prannay Khosla, Piotr Teterwak, Chen Wang, Aaron Sarna, Yonglong Tian, Phillip
  Isola, Aaron Maschinot, Ce~Liu, and Dilip Krishnan.
\newblock Supervised contrastive learning.
\newblock \emph{arXiv preprint arXiv:2004.11362}, 2020.

\bibitem[Kingma and Ba(2014)]{kingma2014adam}
Diederik~P Kingma and Jimmy Ba.
\newblock Adam: A method for stochastic optimization.
\newblock \emph{arXiv preprint arXiv:1412.6980}, 2014.

\bibitem[Kiyasseh et~al.(2020)Kiyasseh, Zhu, and Clifton]{kiyasseh2020clocs}
Dani Kiyasseh, Tingting Zhu, and David~A Clifton.
\newblock Clocs: Contrastive learning of cardiac signals.
\newblock \emph{arXiv preprint arXiv:2005.13249}, 2020.

\bibitem[Kostas et~al.(2021)Kostas, Aroca-Ouellette, and
  Rudzicz]{kostas2021bendr}
Demetres Kostas, Stephane Aroca-Ouellette, and Frank Rudzicz.
\newblock Bendr: using transformers and a contrastive self-supervised learning
  task to learn from massive amounts of eeg data.
\newblock \emph{arXiv preprint arXiv:2101.12037}, 2021.

\bibitem[Kuperman et~al.(2007)Kuperman, Bobb, Payne, Avery, Gandhi, Burns,
  Classen, and Bates]{kuperman2007medication}
Gilad~J Kuperman, Anne Bobb, Thomas~H Payne, Anthony~J Avery, Tejal~K Gandhi,
  Gerard Burns, David~C Classen, and David~W Bates.
\newblock Medication-related clinical decision support in computerized provider
  order entry systems: a review.
\newblock \emph{Journal of the American Medical Informatics Association},
  14\penalty0 (1):\penalty0 29--40, 2007.

\bibitem[Lewis et~al.(2005)Lewis, Foltynie, Blackwell, Robbins, Owen, and
  Barker]{Lewis343}
S~J~G Lewis, T~Foltynie, A~D Blackwell, T~W Robbins, A~M Owen, and R~A Barker.
\newblock Heterogeneity of parkinson{\textquoteright}s disease in the early
  clinical stages using a data driven approach.
\newblock \emph{Journal of Neurology, Neurosurgery \& Psychiatry}, 76\penalty0
  (3):\penalty0 343--348, 2005.

\bibitem[Li et~al.(2020)Li, Li, Zhang, Peng, Zhou, and Gao]{li2020self}
Chunyuan Li, Xiujun Li, Lei Zhang, Baolin Peng, Mingyuan Zhou, and Jianfeng
  Gao.
\newblock Self-supervised pre-training with hard examples improves visual
  representations.
\newblock \emph{arXiv preprint arXiv:2012.13493}, 2020.

\bibitem[Li et~al.(2019)Li, Roberts, Jiang, and Long]{li2019distributed}
Ziyi Li, Kirk Roberts, Xiaoqian Jiang, and Qi~Long.
\newblock Distributed learning from multiple ehr databases: contextual
  embedding models for medical events.
\newblock \emph{Journal of biomedical informatics}, 92:\penalty0 103138, 2019.

\bibitem[Lin et~al.(2017)Lin, Goyal, Girshick, He, and
  Doll{\'a}r]{lin2017focal}
Tsung-Yi Lin, Priya Goyal, Ross Girshick, Kaiming He, and Piotr Doll{\'a}r.
\newblock Focal loss for dense object detection.
\newblock In \emph{Proceedings of the IEEE international conference on computer
  vision}, pages 2980--2988, 2017.

\bibitem[Oord et~al.(2018)Oord, Li, and Vinyals]{oord2018representation}
Aaron van~den Oord, Yazhe Li, and Oriol Vinyals.
\newblock Representation learning with contrastive predictive coding.
\newblock \emph{arXiv preprint arXiv:1807.03748}, 2018.

\bibitem[Santiso et~al.(2019)Santiso, Casillas, and
  P{\'e}rez]{santiso2019class}
Sara Santiso, Arantza Casillas, and Alicia P{\'e}rez.
\newblock The class imbalance problem detecting adverse drug reactions in
  electronic health records.
\newblock \emph{Health informatics journal}, 25\penalty0 (4):\penalty0
  1768--1778, 2019.

\bibitem[Schulam et~al.(2015)Schulam, Wigley, and Saria]{schulam2015clustering}
Peter Schulam, Fredrick Wigley, and Suchi Saria.
\newblock Clustering longitudinal clinical marker trajectories from electronic
  health data: Applications to phenotyping and endotype discovery.
\newblock In \emph{Proceedings of the AAAI Conference on Artificial
  Intelligence}, volume~29, 2015.

\bibitem[Shickel et~al.(2017)Shickel, Tighe, Bihorac, and
  Rashidi]{shickel2017deep}
Benjamin Shickel, Patrick~James Tighe, Azra Bihorac, and Parisa Rashidi.
\newblock Deep ehr: a survey of recent advances in deep learning techniques for
  electronic health record (ehr) analysis.
\newblock \emph{IEEE journal of biomedical and health informatics}, 22\penalty0
  (5):\penalty0 1589--1604, 2017.

\bibitem[Sohn(2016)]{sohn2016improved}
Kihyuk Sohn.
\newblock Improved deep metric learning with multi-class n-pair loss objective.
\newblock In \emph{Proceedings of the 30th International Conference on Neural
  Information Processing Systems}, pages 1857--1865, 2016.

\bibitem[Su et~al.(2021)Su, Zhang, Flory, Weiner, Kaushal, Schenck, and
  Wang]{su2021novel}
Chang Su, Yongkang Zhang, James~H Flory, Mark~G Weiner, Rainu Kaushal, Edward~J
  Schenck, and Fei Wang.
\newblock Novel clinical subphenotypes in covid-19: derivation, validation,
  prediction, temporal patterns, and interaction with social determinants of
  health.
\newblock \emph{medRxiv}, 2021.

\bibitem[Suresh et~al.(2018)Suresh, Gong, and Guttag]{suresh2018learning}
Harini Suresh, Jen~J Gong, and John~V Guttag.
\newblock Learning tasks for multitask learning: Heterogenous patient
  populations in the icu.
\newblock In \emph{Proceedings of the 24th ACM SIGKDD International Conference
  on Knowledge Discovery \& Data Mining}, pages 802--810, 2018.

\bibitem[Verhaak et~al.(2010)Verhaak, Hoadley, Purdom, Wang, Qi, Wilkerson,
  Miller, Ding, Golub, Mesirov, et~al.]{verhaak2010integrated}
Roel~GW Verhaak, Katherine~A Hoadley, Elizabeth Purdom, Victoria Wang, Yuan Qi,
  Matthew~D Wilkerson, C~Ryan Miller, Li~Ding, Todd Golub, Jill~P Mesirov,
  et~al.
\newblock Integrated genomic analysis identifies clinically relevant subtypes
  of glioblastoma characterized by abnormalities in pdgfra, idh1, egfr, and
  nf1.
\newblock \emph{Cancer cell}, 17\penalty0 (1):\penalty0 98--110, 2010.

\bibitem[Wang et~al.(2018)Wang, Xu, and Li]{wang2018utility}
Liansheng Wang, Qiuhao Xu, and Shuo Li.
\newblock Utility balanced classification for automatic electronic medical
  record analysis.
\newblock In \emph{2018 5th International Conference on Systems and Informatics
  (ICSAI)}, pages 1093--1098. IEEE, 2018.

\bibitem[Wang et~al.(2020)Wang, Zhu, Li, Yin, and Zhang]{wang2020feature}
Zhe Wang, Yiwen Zhu, Dongdong Li, Yichao Yin, and Jing Zhang.
\newblock Feature rearrangement based deep learning system for predicting heart
  failure mortality.
\newblock \emph{Computer methods and programs in biomedicine}, 191:\penalty0
  105383, 2020.

\bibitem[Wanyan et~al.(2021)Wanyan, Honarvar, Jaladanki, Zang, Naik, Somani,
  De~Freitas, Paranjpe, Vaid, Miotto, et~al.]{wanyan2021contrastive}
Tingyi Wanyan, Hossein Honarvar, Suraj~K Jaladanki, Chengxi Zang, Nidhi Naik,
  Sulaiman Somani, Jessica~K De~Freitas, Ishan Paranjpe, Akhil Vaid, Riccardo
  Miotto, et~al.
\newblock Contrastive learning improves critical event prediction in covid-19
  patients.
\newblock \emph{arXiv preprint arXiv:2101.04013}, 2021.

\bibitem[Weinberger and Saul(2009)]{weinberger2009distance}
Kilian~Q Weinberger and Lawrence~K Saul.
\newblock Distance metric learning for large margin nearest neighbor
  classification.
\newblock \emph{Journal of machine learning research}, 10\penalty0 (2), 2009.

\bibitem[Wu et~al.(2010)Wu, Roy, and Stewart]{wu2010prediction}
Jionglin Wu, Jason Roy, and Walter~F Stewart.
\newblock Prediction modeling using ehr data: challenges, strategies, and a
  comparison of machine learning approaches.
\newblock \emph{Medical care}, pages S106--S113, 2010.

\bibitem[Yang and Xu(2020)]{yang2020rethinking}
Yuzhe Yang and Zhi Xu.
\newblock Rethinking the value of labels for improving class-imbalanced
  learning.
\newblock In \emph{Proceedings of the 34th Conference on Neural Information
  Processing Systems (NeurIPS 2020)}, 2020.

\end{thebibliography}
\newpage
\appendix
\section*{Appendix A. Input Features For EHR data}
For our experiment, the collected EHR data contains the following specific features: COVID-19 status and demographics (age, gender, and race). We also collect vital signs, specifically: heart rate, respiration rate, pulse oximetry, blood pressure (diastolic and systolic), temperature, height, and weight. We compile 12 comorbidities: alcoholism, asthma, atrial fibrillation, coronary artery disease, cancer, chronic kidney disease, chronic obstructive pulmonary disease, diabetes mellitus, heart failure, hypertension, stroke, and liver disease. Lastly, we collect 34 relevant laboratory test results, these lab tests are specifically listed as follows:\\
Albumin, Alkaline Phosphatase, Alanine Aminotransferase, Amylase, Anion Gap, Partial Thromboplastin Time, Aspartate transaminase, Atypical Lymphocytes, Band Cells, Basophil \%, Basophil Count, Direct Bilirubin, Total Bilirubin, Myeloblasts, Brain Natriuretic Peptide, Blood Urea Nitrogen, C-reactive protein, Ionized Calcium, Calcium, Chloride, Creatine phosphokinase, Creatine kinase-MB, Bicarbonate, Creatinine, D-dimer, Eosinophil \%, Eosinophil, Ferritin, Fibrogen, Glucose, Hematocrit, Hemoglobin, International Normalized Ratio, IL6, Iron, Ketone, Lactate, Lactate Dehydrogenase, Lymphocyte \%, Lymphocytes, Mean Corpuscular Hemoglobin Concentration, Mean Corpuscular Volume, Mean Platelet Volume, Monocyte \%, Monocytes, Neutrophil \%, Neutrophils, Oxygen Saturation, pH, Platelets, Partial Pressure of Oxygen, Potassium, PT, Protein, Red Blood Cell Count, Red Cell Distribution Width, Sodium, Total Iron-Binding Capacity, Transferrin saturation, Troponin I, Uric Acid, White Blood Cell Count

\section*{Appendix B. Studying the Effect of Model Parameters for Contrastive Regularizer}
We assess the effect of different parameters for our three positive sampling contrastive regularizer models. There are three model parameters that could be customized and optimized in our contrastive learning framework, specifically: $k,\alpha,\tau$, where $k$ is the positive sample numbers, $\alpha$ is the weight for contrastive regularizer, $\tau$ is the temperature parameter for contrastive loss. First, we test on the performance using different k values, with $\alpha=0.2$ and $\tau=1$ fixed. the result is shown in Table~\ref{tab:k}. 

\begin{table}[h]
  \centering
  \caption{Prediction Performance on Different k}
    \begin{tabular}{p{3cm}p{1.5cm}p{1.5cm}p{1.5cm}p{1.5cm}p{1.5cm}p{1.5cm}}
    \toprule
    Loss Model &k=1 & k=3 & k=5  &k=9 &k=11  &k=15\\
    \midrule
    FL(random) &0.949 &0.951 &0.950 &0.953 &0.952 &0.953\\
    FL(feature) &0.950 &0.957 &0.956 &0.957 &0.953 &0.955\\
    FL(attribute)&0.956 &0.957 &0.954 &0.961 &0.956 &0.961\\
    \midrule
    \end{tabular}%
  \label{tab:k}%
\end{table}

Second, we test the performance on the impact of different regularizer weight coefficient $\alpha$ with $k=5$ and $\tau=1$ fixed. The results of this experiment are shown in Table~\ref{tab:alpha}
\begin{table}[t]
  \centering
  \caption{Predictive Performance using Different Weight Coefficients $\alpha$}
    \begin{tabular}{p{3cm}p{2cm}p{2cm}p{2cm}p{2cm}p{2cm}}
    \toprule
    Loss Model &$\alpha=0.1$ &$\alpha=0.2$ & $\alpha=0.4$ &$\alpha=0.6$ &$\alpha=0.8$\\
    \midrule
    FL(random) &0.946 &0.949 &0.953 &0.955 &0.956\\
    FL(feature) &0.952 &0.953 &0.956 &0.955 &0.955\\
    FL(attribute)&0.951 &0.954 &0.958 &0.958 &0.957\\
    \midrule
    \end{tabular}%
  \label{tab:alpha}%
\end{table}

Finally, we test the parameter $\tau$ with $k=5$ and $\alpha=0.2$ fixed. The results are shown in Table~\ref{tab:tau}.
\begin{table}[t]
  \centering
  \caption{Predictive Performance using Different $\tau$}
    \begin{tabular}{p{3cm}p{2cm}p{2cm}p{2cm}p{2cm}}
    \toprule
    Loss Model &$\tau=0.1$ &$\tau=0.5$ & $\tau=1$ &$\tau=1.5$ \\
    \midrule
    FL(random) &0.961 &0.962 &0.953 &0.954 \\
    FL(feature) &0.963 &0.965 &0.956 &0.955 \\
    FL(attribute)&0.967 &0.966 &0.958 &0.957 \\
    \midrule
    \end{tabular}%
  \label{tab:tau}%
\end{table}

\end{document}